\documentclass[10pt,conference]{IEEEtran}
\IEEEoverridecommandlockouts

\usepackage{etoolbox}
\makeatletter
\patchcmd{\@makecaption}
  {\scshape}
  {}
  {}
  {}
\makeatletter
\patchcmd{\@makecaption}
  {\\}
  {.\ }
  {}
  {}
\makeatother

\usepackage{cite}
\usepackage{amsmath,amssymb,amsfonts}
\usepackage{algorithmic}
\usepackage{graphicx}
\usepackage{textcomp}
\usepackage{xcolor}
\def\BibTeX{{\rm B\kern-.05em{\sc i\kern-.025em b}\kern-.08em
    T\kern-.1667em\lower.7ex\hbox{E}\kern-.125emX}}

\usepackage{colortbl} 
\usepackage{pifont}
\usepackage{booktabs}
\usepackage{multirow}
\usepackage{algorithm}
\usepackage{algorithmic}
\usepackage{hyperref}
\begin{document}

\title{CodeACT: Code Adaptive Compute-efficient Tuning Framework for Code LLMs}

\author{\IEEEauthorblockN{Weijie Lv}
\IEEEauthorblockA{
\textit{Nanjing University of}\\
\textit{Aeronautics and Astronautics}\\
Nanjing, China \\
lvweijie@nuaa.edu.cn}
\and
\IEEEauthorblockN{Xuan Xia}
\IEEEauthorblockA{
\textit{Shenzhen Institute of Artificial}\\
\textit{Intelligence and Robotics for Society}\\
Shenzhen, China \\
xiaxuan@cuhk.edu.cn}
\and
\IEEEauthorblockN{Sheng-Jun Huang}
\IEEEauthorblockA{
\textit{Nanjing University of}\\
\textit{Aeronautics and Astronautics}\\
Nanjing, China \\
huangsj@nuaa.edu.cn}
}

\maketitle

\begin{abstract}
Large language models (LLMs) have shown great potential in code-related tasks, yet open-source models lag behind their closed-source counterparts. To bridge this performance gap, existing methods generate vast amounts of synthetic data for fine-tuning, leading to inefficiencies in training. Motivated by the need for more effective and efficient training, we propose the Code Adaptive Compute-efficient Tuning (CodeACT) framework. CodeACT introduces the Complexity and Diversity Aware Sampling (CDAS) method to select high-quality training data based on complexity and diversity, and the Dynamic Pack padding strategy to reduce computational resource usage by minimizing padding tokens during training. Experimental results demonstrate that CodeACT-DeepSeek-Coder-6.7B, fine-tuned on only 40\% of the EVOL-Instruct data, achieves an 8.6\% performance increase on HumanEval, reduces training time by 78\%, and decreases peak GPU memory usage by 27\%. These findings underscore CodeACT's ability to enhance the performance and efficiency of open-source models. By optimizing both the data selection and training processes, CodeACT offers a comprehensive approach to improving the capabilities of open-source LLMs while significantly reducing computational requirements, addressing the dual challenges of data quality and training efficiency, and paving the way for more resource-efficient and performant models. The code is available at \href{https://github.com/Kyle-Lyu/CodeACT}{Kyle-Lyu/CodeACT}.
\end{abstract}

\begin{IEEEkeywords}
AI4SE, Large Language Models, Code Generation, Data Selection, Compute-efficient Tuning
\end{IEEEkeywords}

\section{Introduction}

Large language models (LLMs) have recently achieved remarkable success across various domains, with their applications in code-related tasks emerging as a focal point in software engineering research. Pioneered by models like Codex\cite{codex}, LLMs have demonstrated exceptional prowess in code processing, leading to the development of commercial products such as GitHub Copilot\footnote{https://github.com/features/copilot} and open-source alternatives like CodeLlama\cite{codellama}, DeepSeek-Coder\cite{deepseek-coder}, and StarCoder\cite{starcoder2}. However, a persistent performance gap remains between open-source and closed-source models, particularly in code generation tasks.

Instruction fine-tuning is a method employed to refine the performance of LLMs, predominantly relying on amassing large datasets. Approaches such as CodeAlpaca\cite{codealpaca}, EVOL-Instruct\cite{wizardcoder}, and OSS-Instruct\cite{magicoder} have leveraged more powerful LLMs (e.g., GPT-4\cite{gpt-4}) to generate synthetic coding instructions and fine-tune open-source models. While these methods have shown promise, they often lead to inefficient optimization and training processes due to the presence of low-quality synthetic data within the massive training corpora.

Recent research, exemplified by LIMA\cite{lima}, suggests that data quality is more crucial than quantity, demonstrating superior performance with just 1,000 carefully curated samples. This insight raises a critical question: \textit{How can we identify the most influential training samples to enhance model performance and training efficiency simultaneously?}

Complex programming problems typically require an integration of various knowledge domains and skills, demanding more intricate reasoning processes than simpler ones. Intuitively, these complex problems could contribute substantially to model training. Additionally, numerous studies\cite{DEITA, qdit, qads, car},  have highlighted the importance of data diversity in improving model performance. These observations suggest that selecting diverse and complex code data could be key to efficient and effective model training.

Building upon these insights, we propose a novel framework called \textbf{Code} \textbf{A}daptive \textbf{C}ompute-efficient \textbf{T}uning (\textbf{CodeACT}). This framework addresses both the quality of training data and the efficiency of the fine-tuning process, two interrelated aspects that collectively impact the performance and resource utilization of Code LLMs. At the core of CodeACT is the \textbf{C}omplexity and \textbf{D}iversity \textbf{A}ware \textbf{S}ampling (\textbf{CDAS}) method, specifically designed to identify the most influential code data. Notably, CDAS operates adaptively by utilizing the base LLM for data selection, thereby eliminating the need for external LLMs. By selecting a smaller set of high-quality data, CDAS aims to enhance training efficiency while maintaining or improving model performance.

Complementing CDAS, we introduce the \textbf{Dynamic Pack} padding strategy to optimize the resource utilization. Traditional padding strategies often introduce a large number of padding tokens, leading to inefficient resource utilization and prolonged training times. Dynamic Pack addresses this by sorting data within a batch by length and merging multiple instances, effectively reducing the rate of padding tokens. This technique not only accelerates the training process but also decreases computational resource consumption, further enhancing the efficiency gains achieved through CDAS's data selection.

The synergy between CDAS and Dynamic Pack in the CodeACT framework enables significant improvements in both model performance and training efficiency. CDAS contributes to efficiency by selecting a reduced set of influential data, while Dynamic Pack further boosts efficiency by minimizing padding tokens during training. Our experiments demonstrate that CodeACT-DeepSeek-Coder-7B, fine-tuned on only 40\% of the EVOL-Instruct data, achieves an 8.6\% increase on HumanEval (from 58.5\% to 67.1\%) compared to training with the full dataset. Moreover, it reduces training time by 78\% (from 297 to 68 minutes) and decreases peak GPU memory usage by 27\% (from 50.75 GB to 37.23 GB).

The main contributions of this paper are as follows:
\begin{itemize}
\item We propose the CodeACT framework, which integrates data selection and an efficient padding strategy to enhance both the performance and training efficiency of LLMs.
\item We introduce the CDAS method, an adaptive sampling method specifically designed for code data, which identifies influential training samples by considering both complexity and diversity.
\item We develop the Dynamic Pack strategy, which significantly reduces padding tokens during the training phase, thereby further improving training efficiency and resource utilization.
\item Extensive experimental results validate the effectiveness of our framework, demonstrating superior performance with less data and substantially enhanced training efficiency.
\end{itemize}

\section{Preliminaries}

We denote a dataset as $D$, which consists of $n$ triplets $x = (Instruction, [Input], Response)$ representing instruction tuning data samples. Earlier instruction tuning samples typically feature separate \textit{instruction} and \textit{input} segments for better control \cite{wang-etal-2022-super, longpre2023flan, alpaca}, while most current datasets integrate the inputs with instructions \cite{lima, vicuna2023, xu2024wizardlm, li2023reflection}. For simplicity, let $q = map(Instruction, [Input])$ denote the complete instruction and $a$ as the corresponding response. The mapping function may simply concatenate them with control tokens. Thus, $D = \{(q_1, a_1), (q_2, a_2), \ldots, (q_n, a_n)\}$ represents a collection of $n$ instruction-response pairs.

\subsection{Perplexity}

In the context of instruction tuning, the objective is to maximize the likelihood of generating the correct response given the corresponding instruction. Therefore, perplexity can serve as a potential metric for assessing the difficulty of samples. Specifically, the perplexity of a given sample \( (q_i, a_i) \) is defined as:
\begin{equation}
    \text{PPL}(a_i \mid q_i) = e^{ -\frac{1}{N} \sum_{j=1}^{N} \log P(a_{i,j} \mid q_i, a_{i,1}, \ldots, a_{i,j-1}) },
\label{eq:1}
\end{equation}
where \( N \) is the length of the response \( a_i \), and \( a_{i,j} \) represents the \( j \)-th token in the response \( a_i \).

\subsection{Instruction-Following Difficulty Score} 

Cherry LLM\cite{ifd} introduces a self-guided approach for evaluating data complexity that does not rely on powerful external models (e.g., GPT-4). This method leverages the Instruction-Following Difficulty (IFD) score, which is calculated using an originally pre-trained LLM. The IFD score is a purely statistical measure that compares the losses or perplexities when the model generates a response \( a \) with and without the instructional context \( q \). This comparison quantifies the extent to which the instruction aids in generating the corresponding response. This approach is particularly advantageous as it allows the model to autonomously evaluate the complexity of the data, thereby making the process more efficient and scalable.

A higher IFD score indicates that, even with the given instruction $q$, the model struggles to generate an accurate response $a$, reflecting the complexity of the sample. Conversely, a lower IFD score suggests that the instruction $q$ significantly facilitates the generation of the correct response \( a \) without needing further training, which may be due to the sample being straightforward or the instruction containing detailed information that aligns closely with the model's pre-trained knowledge. The IFD score for a given instruction-response data pair is calculated as follows:
\begin{equation}
    \text{IFD}(a_i \mid q_i) = \frac{\text{PPL}(a_i \mid q_i)}{\text{PPL}(a_i)},
\label{eq:2}
\end{equation}
where \(\text{PPL}(a_i \mid q_i)\) and \(\text{PPL}(a_i)\) denote the perplexities of the model in fitting response \( a_i \) with and without the instruction \( q_i \), respectively.

\section{Approach}

\subsection{Definitions}

\textbf{Data Selection Task.} Given a vast pool of instruction tuning data $D = \{x_1, x_2, \ldots, x_n\}$, where each $x_i$ represents an individual instruction-response pair ($q_i$, $a_i$), our objective is to select a subset $S_\pi^{(m)}$ of size $m$ from $D$, employing a selection strategy $\pi$. To evaluate the effectiveness of the selected subset, we represent the alignment performance after instruction tuning as $Q$\cite{DEITA}. The optimal selection strategy $\pi^*$  within a given data budget $m$ is then defined as:
\begin{equation}
\pi^* = \arg\max_\pi Q(S_\pi^{(m)}).
\label{eq:3}
\end{equation}

In general, the alignment performance, as indicated by Q, is determined by the test data. When the test data remains consistent, the optimal selection strategy $\pi^*$ should be developed based on quantifiable indicators within the training data. We propose that data complexity and dataset diversity serve as these indicators and aim to quantify them. 

\begin{figure*}
    \centering
    \includegraphics[width=0.85\textwidth]{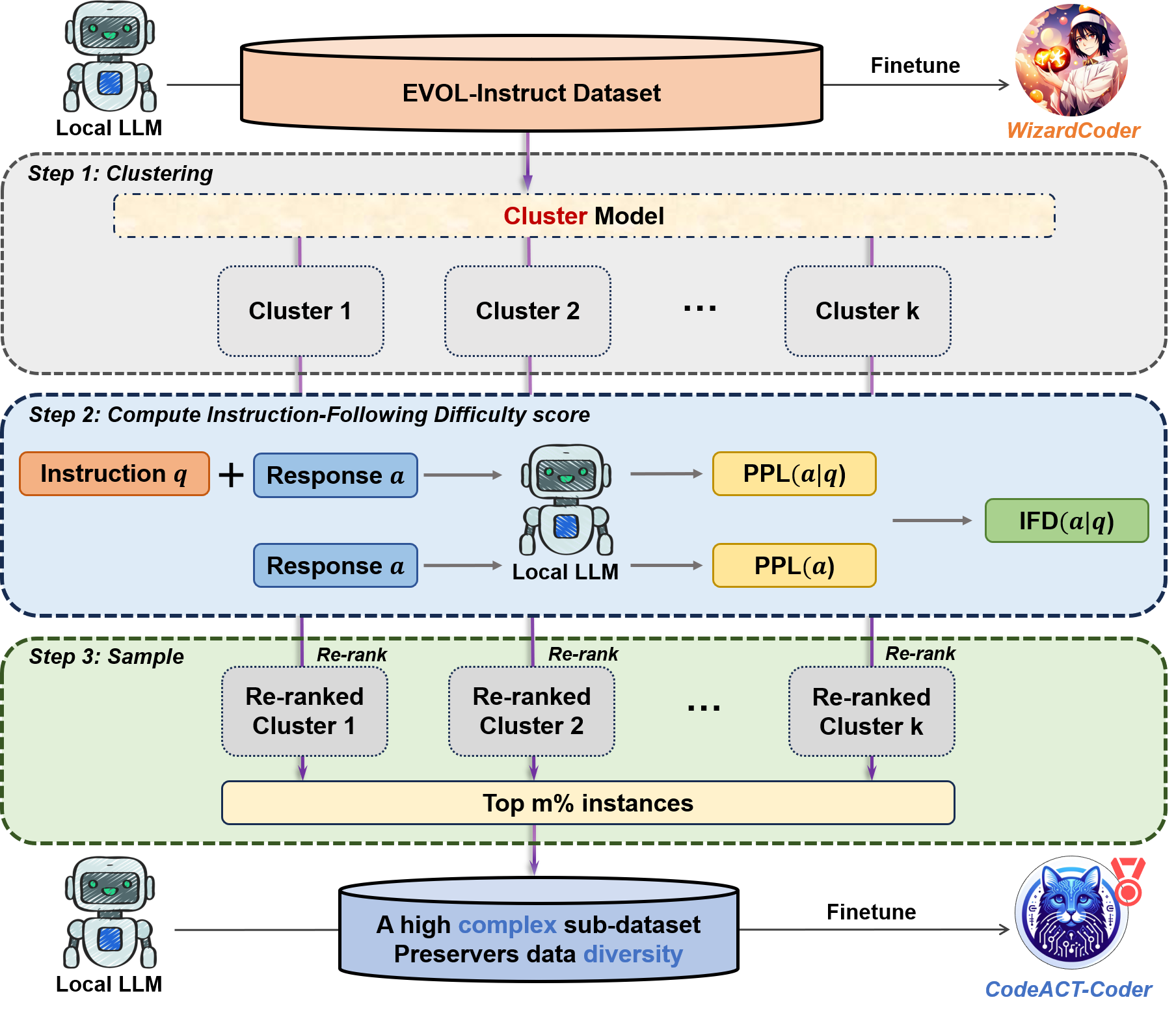}
    \caption{An overviw of our proposed CDAS method, including three steps from top to bottom. Step 1: Clustering the EVOL-Instruct dataset to form multiple clusters. Step 2: Computing the Instruction-Following Difficulty score by comparing the model's perplexity with and without instructions. Step 3: Sampling the top m\% instances from each re-ranked cluster to form a high-complexity sub-dataset that preserves data diversity. Finally, we use the selected data for fine-tuning to obtain CodeACT-Coder.}
    \label{fig:CDAS}
\end{figure*}

\textbf{Complexity.} Complexity measures the difficulty of a single instruction-response pair, where a higher complexity signifies a greater learning value. Complex programming problems typically require the integration of multiple domains of knowledge and skills, necessitating more sophisticated reasoning and a higher level of detail than simpler problems. We introduce the IFD score as a measure of complexity. The IFD score quantifies the extent to which an instruction facilitates the generation of a corresponding response by comparing the model's loss or perplexity. A higher IFD score indicates that the sample likely involves more complex knowledge or uncommon combinations, thereby revealing the problem's intricacy. Conversely, a lower IFD score suggests that the sample is simpler and contains information highly consistent with the model's pre-trained knowledge. 

More importantly, the IFD score serves as a purely statistical measure that can reflect different models' performance on the same dataset. Different models will yield different IFD score distributions, indicating their varied definitions and understandings of complex data. Therefore, IFD demonstrates the model's adaptability, enabling an objective evaluation of their performance across tasks of varying complexity. This characteristic makes the IFD score an effective tool for systematically identifying and selecting complex programming problems.

\textbf{Diversity.} Diversity measures the richness of the entire dataset's samples, with higher diversity indicating a more varied and comprehensive dataset. While complexity is crucial, ensuring data diversity is equally important for enabling the model to perform well across various scenarios \cite{DEITA, qdit, qads, car}. In this paper, we define diversity as the range occupied by the probability distribution of all samples in the dataset $D$ within the semantic space. Although this distribution is unknown and prevents direct calculation of diversity, the dataset's diversity is fixed once created. Thus, in the task of data selection, we aim to maintain the subset's diversity consistent with the original dataset's, as represented by the formula:
\begin{equation}
\lim_{|S| \rightarrow \infty} P_S(x)=P_D(x),
  \label{eq:4}
\end{equation}
where $|S|$ denotes the size of subset $S$, $P_S(x)$ and $P_D(x)$ are the probability distributions of $S$ and $D$, respectively.

The most straightforward approach to satisfy \eqref{eq:4} is through random sampling. Nonetheless, \eqref{eq:4} imposes a weak constraint on subset sampling, becoming effective only when $|S|$ approaches infinity. For smaller $|S|$, it fails to guarantee that the distribution of $S$ in the semantic space aligns with that of $D$. A more effective strategy involves partitioning $D$ into multiple sub-datasets $D_i$, from which we sample subsets $S_i$, ensuring $S$ adheres to:
\begin{equation}
\lim_{|S_i| \rightarrow \infty} P_{S_i}(x)=P_{D_i}(x) \quad \text{for} \quad i = 1, 2, \ldots, k,
  \label{eq:5}
\end{equation}
where $|S_i|$ represents the size of $S_i$, $P_{S_i}(x)$ and $P_{D_i}(x)$ are the probability distributions of $S_i$ and $D_i$, respectively.

Equation \eqref{eq:5} provides stronger constraints compared to \eqref{eq:4}, ensuring that the distribution range of $S$ in semantic space is more consistent with $D$ when the number of samples is reduced. Based on the above analysis, we propose using K-Means to partition the dataset $D$ into sub-datasets $D_i$ to ensure that the diversity of the sampled subset $S$ is consistent with that of the original dataset $D$. To validate our choice of K-Means, we conducted a comparative study with other methods in \textit{Section \ref{subsec:RQ3}}.

\subsection{Complexity and Diversity Aware Sampling}

To address the challenges in selecting optimal code data, we propose the \textbf{C}omplexity and \textbf{D}iversity \textbf{A}ware \textbf{S}ampling (\textbf{CDAS}) method. This approach is rooted in the necessity of considering both the complexity and diversity of data to enhance model training efficiency and effectiveness. By integrating these two aspects, CDAS aims to improve the generalization capabilities of LLMs in the programming domain.

\begin{algorithm}
\caption{Complexity and Diversity Aware Sampling}\label{alg:cdas}
\begin{algorithmic}[1]
\REQUIRE sampling proportion $m\%$, the whole dataset $D = \{(q_1, a_1), (q_2, a_2), \ldots, (q_n, a_n)\}$
\ENSURE Sampled dataset $S$
\STATE Derive embeddings from instructions in $D$
\STATE Partition $D$ into $k$ clusters based on embeddings
\FOR{each cluster $C_k$}
    \FOR{each instance $(q_i, a_i) \in C_k$}
        \STATE Compute IFD score for instance $(q_i, a_i)$
    \ENDFOR
    \STATE Reorder instances in $C_k$ based on IFD scores
    \STATE Sample the top $m\%$ of instances from $C_k$
\ENDFOR
\STATE Combine sampled instances from all clusters to form $S$
\RETURN $S$
\end{algorithmic}
\end{algorithm}

As shown in Algorithm \ref{alg:cdas}, CDAS begins by utilizing lightweight sentence transformers\cite{reimers-2019-sentence-bert} to derive embeddings from the given instructions. These embeddings effectively capture the semantic essence of the instructions, facilitating more accurate clustering. Subsequently, the dataset is partitioned into distinct clusters, which ensures data diversity by aggregating similar instances. Within each cluster, the IFD score is computed for each instance, with the aim of identifying complex programming data. After that, the data within each cluster is reordered, and the top $m\%$ of instances from each cluster is sampled. This final sampling step guarantees that the selected data maintains a balanced representation of complexity and diversity, thereby enhancing the model's ability to generalize across various coding tasks. 

By considering both diversity and complexity dimensions of data, CDAS offers a promising solution to the challenge of selecting optimal instruction tuning data for LLMs in the programming domain. This method paves the way for more effective and efficient model training, potentially leading to significant improvements in code generation and understanding capabilities.

\begin{figure*}
    \centering
    \includegraphics[width=\textwidth]{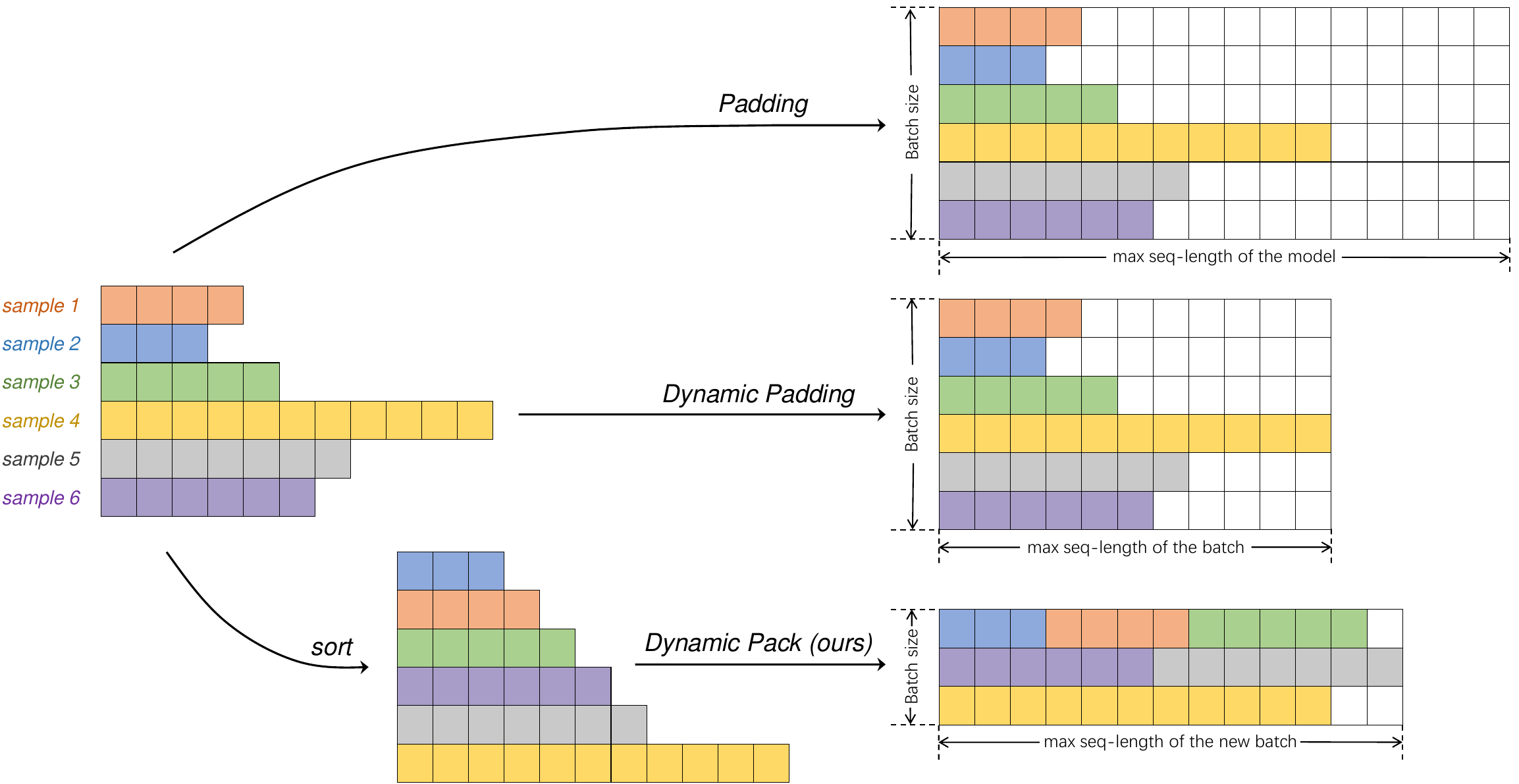}
    \caption{Illustration of different padding strategies, where the blank squares represent padding tokens. Top: Traditional padding strategy aligns samples to the model's maximum input length, resulting in high computational resource consumption. Middle: Dynamic padding strategy reduces the number of padding tokens by aligning samples to the length of the longest sample in each batch. Bottom: Our proposed Dynamic Pack strategy sorts samples by length and concatenates multiple samples within a batch, further optimizing the utilization of the model's maximum input length and reducing padding tokens.}
    \label{fig:DynamicPack}
\end{figure*}

\subsection{Padding Strategy}

Tokenization is a crucial step in the pre-training and fine-tuning processes of LLMs. The primary task of tokenization is to segment the text into smaller units for processing. This step not only effectively defines how data is utilized during training but also plays a pivotal role in enhancing the model's effectiveness and training efficiency. However, due to varying sample lengths, traditional padding strategy typically aligns samples to the model's maximum input length by using additional padding tokens, as shown at the top of Figure \ref{fig:DynamicPack}. This often results in a high proportion of padding tokens, which reduces training efficiency and increases computational resource consumption.

To address this issue, the dynamic padding strategy has been proposed and widely adopted. In this strategy, the maximum input length is determined by the longest sample in each batch, with shorter samples being padded to match this length, as illustrated in the middle of Figure \ref{fig:DynamicPack}. It effectively reduces the number of padding tokens used, thereby significantly accelerating the training process.

To further optimize the utilization of maximum input length and reduce the number of padding tokens, we propose the \textbf{Dynamic Pack} strategy. This strategy first sorts the samples within a batch by length and then attempts to concatenate multiple samples into a single data instance, as shown at the bottom of Figure \ref{fig:DynamicPack}. This process results in a new batch of samples, which are then padded based on the maximum length of the new batch. The Dynamic Pack strategy not only enhances training efficiency but also further optimizes the use of computational resources.

\subsection{Code Adaptive Compute-efficient Tuning Framework}

The \textbf{Code} \textbf{A}daptive \textbf{C}ompute-efficient \textbf{T}uning (\textbf{CodeACT}) framework is meticulously crafted to refine the training process of LLMs through the integration of a sophisticated data selection method and an innovative token padding strategy. The overarching objective of CodeACT is to amplify the efficiency and efficacy of model training, particularly for intricate tasks such as code generation.

CodeACT harnesses the \textbf{CDAS} method for code data selection. CDAS ensures that the selected data is both diverse and complex, thereby facilitating the creation of a robust training dataset that bolsters the model's generalization capabilities. Additionally, CodeACT introduces the \textbf{Dynamic Pack} padding strategy. This innovative strategy overcomes the inefficiencies of traditional and dynamic padding by sorting samples by length and concatenating them without surpassing the model's maximum input length. The Dynamic Pack strategy significantly reduces the number of padding tokens, optimizing computational resources and expediting the training process.

By seamlessly integrating CDAS and Dynamic Pack, CodeACT offers a holistic solution for enhancing the training efficiency and performance of LLMs in complex tasks. This framework not only elevates model performance in code generation but also lays the foundation for more effective and resource-efficient training processes.

\section{Experimental Setup}

\subsection{Datasets}

\textbf{EVOL-Instruct.} The EVOL-Instruct\cite{wizardcoder} dataset is derived from the iterative evolution of the Code Alpaca\cite{codealpaca} dataset, where instruction complexity is incrementally increased using ChatGPT with evolution prompts. These prompts encompass five distinct aspects, including the imposition of constraints, the substitution of broad requirements with more detailed ones, the extension of reasoning steps, the inclusion of deceptive code, and the enhancement of time or space complexity. Each instruction undergoes multiple iterations of evolution, during which pruning and post-processing are performed to eliminate undesirable instructions and responses. This iterative complexity augmentation method produces instructions of higher quality and depth compared to traditional Alpaca methods. We utilize the \textit{Evol-Instruct-Code-80K\footnote{https://huggingface.co/datasets/nickrosh/Evol-Instruct-Code-80k-v1}} dataset, an open-source implementation comprising approximately 80K samples.

\textbf{OSS-Instruct.} The OSS-Instruct\cite{magicoder} dataset leverages ChatGPT to generate programming problems and their corresponding solutions. The generation process is controlled by using real code snippets sourced from open-source codebases like GitHub as seeds. This approach is distinctive because it provides real-world code snippets as inspiration, prompting the language model to generate problems that closely reflect actual programming scenarios. This not only ensures the diversity and authenticity of the generated problems but also captures the various challenges encountered in real-world programming.

\begin{table*}[ht]
\renewcommand{\arraystretch}{1.2}
\centering
\caption{Performance comparison of our framework across different datasets and models. The \textit{CodeACT} column indicates whether the model was trained using our framework. The \textbf{bold} scores represent the best performance achieved using the same base model. The results highlight the efficiency gains achieved by CodeACT in terms of reduced training time and peak GPU memory usage, while maintaining or improving performance across various benchmarks.}
\begin{tabular}{lcccccccc}
\toprule
\multirow{2}{*}{\textbf{Model}} & \multirow{2}{*}{\textbf{Size}} & \multirow{2}{*}{\textbf{CodeACT}} &  \multicolumn{2}{c}{\textbf{Efficiency}}  & \multicolumn{4}{c}{\textbf{Benchmark(Pass@1 \%)}} \\
\cmidrule(lr){4-5} \cmidrule(lr){6-9}
& & & \textbf{Training Time} & \textbf{Peak GPU Memory} & \textbf{HumanEval} & \textbf{HumanEval+} & \textbf{MBPP} & \textbf{MBPP+} \\
\specialrule{0.1em}{1pt}{1pt}
\multicolumn{8}{c}{\textit{Models trained on OSS-Instruct dataset}} \\
\specialrule{0.1em}{1pt}{1pt}
\multirow{2}{*}{CodeLlama} & \multirow{2}{*}{7B} & \ding{55} & 220 min & 64.28 GB & 50.6 & 47.0 & \textbf{63.2} & \textbf{51.4} \\
& & \ding{51} & \textbf{63 min} & \textbf{33.21 GB} & \textbf{54.3} & \textbf{50.0} & 60.4 & 50.4 \\
\specialrule{0.05em}{1pt}{1pt}
\multirow{2}{*}{CodeLlama} & \multirow{2}{*}{13B} & \ding{55} & 367 min & 69.17 GB & 58.5 & \textbf{52.4} & \textbf{63.2} & \textbf{51.9} \\
& & \ding{51} & \textbf{109 min} & \textbf{59.62 GB} & \textbf{59.8} & \textbf{52.4} & \textbf{63.2} & 50.6 \\
\specialrule{0.05em}{1pt}{1pt}
\multirow{2}{*}{DeepSeek-Coder} & \multirow{2}{*}{6.7B} & \ding{55} & 184 min & 64.48 GB & 65.2 & 61.0 & \textbf{75.9} & \textbf{63.4} \\
& & \ding{51} & \textbf{54 min} & \textbf{35.51 GB} & \textbf{68.3} & \textbf{61.6} & \textbf{75.9} & 61.7 \\
\specialrule{0.1em}{1pt}{1pt}
\multicolumn{8}{c}{\textit{Models trained on EVOL-Instruct dataset}} \\
\specialrule{0.1em}{1pt}{1pt}
\multirow{2}{*}{CodeLlama} & \multirow{2}{*}{7B} & \ding{55} & 297 min & 50.65 GB & \textbf{54.3} & \textbf{50.0} & \textbf{60.7} & 48.6 \\
& & \ding{51} & \textbf{68 min} & \textbf{38.25 GB} & 53.0 & 47.0 & \textbf{60.7} & \textbf{49.9} \\
\specialrule{0.05em}{1pt}{1pt}
\multirow{2}{*}{CodeLlama} & \multirow{2}{*}{13B} & \ding{55} & 468 min & 69.17 GB & 62.2 & \textbf{56.7} & \textbf{63.2} & \textbf{52.9} \\
& & \ding{51} & \textbf{116 min} & \textbf{58.82 GB} & \textbf{64.0} & 55.5 & 62.4 & 51.6 \\
\specialrule{0.05em}{1pt}{1pt}
\multirow{2}{*}{DeepSeek-Coder} & \multirow{2}{*}{6.7B} & \ding{55} & 259 min & 50.75 GB & 58.5 & 53.7 & \textbf{71.4} & \textbf{58.1} \\
& & \ding{51} & \textbf{58 min} & \textbf{37.23 GB} & \textbf{67.1} & \textbf{59.8} & 69.9 & \textbf{58.1} \\
\bottomrule
\end{tabular}
\label{tab:codeact_performance}
\end{table*}

\subsection{Benchmarks}

We employ four code benchmarks: HumanEval\cite{humaneval}, HumanEval+\cite{evalplus}, MBPP\cite{mbpp}, and MBPP+\cite{evalplus}. Consistent with previous research\cite{magicoder, evalplus, chen2024teaching}, we use greedy decoding to generate a single sample for each benchmark and LLM, focusing our comparison on the pass@1 metric.

\textbf{HumanEval/HumanEval+.} HumanEval and its enhanced counterpart, HumanEval+, serve as critical benchmarks for evaluating the code generation capabilities of LLMs. HumanEval comprises 164 manually-written Python problems, each accompanied by an average of 9.6 test cases. HumanEval+ builds upon this by significantly increasing the number of test cases through the use of LLMs and mutation strategies, resulting in a more rigorous evaluation framework.

\textbf{MBPP/MBPP+.} The MBPP (Mostly Basic Python Programming) benchmark includes approximately 1,000 Python challenges, crowd-sourced to assess fundamental programming skills and standard library usage. These challenges are geared towards beginners and each provides a description, a solution, and three tests to verify solution accuracy. MBPP+ extends the MBPP benchmark by incorporating a subset of hand-verified problems from MBPP-sanitized dataset, ensuring that the tasks are well-defined and unambiguous, thereby enhancing the benchmark's reliability and applicability in more rigorous evaluations.

\subsection{Implementation Details}

We employ three base models for our experiments, including DeepSeek-Coder-Base-6.7B, CodeLlama-Python-7B, and CodeLlama-Python-13B. All models are fine-tuned for 3 epochs using eight NVIDIA A100-80GB GPUs through the Fully Sharded Data Parallel (FSDP) module within PyTorch.

The training parameters are consistent across all models, with the exception of the maximum input length for the CodeLlama-Python-13B model. Specifically, we use AdamW\cite{adamw} as our optimizer with a learning rate of 5e-5, a Cosine learning rate scheduler, and 15 warmup steps. The maximum sequence length is set to 4096 for both CodeLlama-Python-7B and DeepSeek-Coder-Base-6.7B, while it is set to 2048 for CodeLlama-Python-13B. The global batch size for all experiments is set to 512. We implement a full-parameter tuning approach throughout the training process.

\section{Experimental Results}

In this section, we present and analyze the findings by addressing five specific research questions.

\begin{table*}[ht]
\renewcommand{\arraystretch}{1.2}
\centering
\caption{Performance comparison of models trained with CodeACT to other models. The \textbf{bold} scores indicate the highest performance among models of the same size. The results show that models trained with CodeACT outperform their base models and achieve competitive results compared to other state-of-the-art open-source models. This underscores the effectiveness of the CodeACT framework in optimizing model performance and efficiency.}
\begin{tabular}{lcccccccc}
\toprule
\multirow{2}{*}{\textbf{Model}} & \multirow{2}{*}{\textbf{Size}} & \multirow{2}{*}{\textbf{Base Model}} & \multirow{2}{*}{\textbf{Data Type}} & \multirow{2}{*}{\textbf{Data Num}} & \multicolumn{4}{c}{\textbf{Benchmark(Pass@1 \%)}} \\
\cmidrule(lr){6-9}
& & & & & \textbf{HumanEval} & \textbf{HumanEval+} & \textbf{MBPP} & \textbf{MBPP+} \\
\midrule
\multicolumn{9}{c}{\textit{Closed-source Models}} \\
\midrule
Gemini-Pro-1.0 & - & - & Proprietary & - & 63.4 & 55.5 & 75.4 & 61.4 \\
Claude-3-Opus & - & - & Proprietary & - & 82.9 & 77.4 & \textbf{89.4} & \textbf{73.3} \\
GPT-4-Turbo & - & - & Proprietary & - & \textbf{85.4} & \textbf{81.7} & 85.7 & \textbf{73.3} \\
\midrule
\multicolumn{9}{c}{\textit{Open-source Models}} \\
\midrule
CodeLlama & 34B & Llama2 & Proprietary & - & 51.8 & 43.9 & 65.4 & 52.6 \\
WizardCoder-CL & 34B & CodeLlama & EVOL-Instruct & 78K & \textbf{73.2} & \textbf{64.6} & \textbf{73.2} & \textbf{59.9} \\
\midrule
StarCoder & 15B & StarCoderBase & Proprietary & - & 34.1 & 29.3 & 55.1 & 46.1 \\
CodeLlama & 13B & Llama 2 & Proprietary & - & 43.3 & 36.6 & 57.6 & 46.9 \\
WizardCoder-SC & 15B & StarCoder & EVOL-Instruct & 78K & 56.7 & 50.6 & 59.6 & 48.1 \\
CodeACT-CL (ours) & 13B & CodeLlama & EVOL-Instruct & 31K & \textbf{64.0} & \textbf{55.5} & \textbf{62.4} & \textbf{51.6} \\
\midrule
CodeLlama & 7B & Llama 2 & Proprietary & - & 39.0 & 34.1 & 58.1 & 46.1 \\
DeepSeek-Coder-Base & 6.7B & - & Proprietary & - & 47.6 & 40.2 & 69.2 & 54.6 \\
WizardCoder-CL & 7B & CodeLlama & EVOL-Instruct & 78K & 50.6 & 45.1 & 58.5 & 49.5 \\
\rowcolor{white} CodeACT-CL (ours) & 7B & CodeLlama & EVOL-Instruct & 31K & 53.0 & 47.0 & 60.7 & 49.9 \\
Magicoder-DS & 6.7B & DeepSeek-Coder & OSS-Instruct & 75K & 66.5 & 60.4 & 75.4 & \textbf{61.9} \\
\rowcolor{white} CodeACT-DS (ours) & 6.7B & DeepSeek-Coder & OSS-Instruct & 30K & \textbf{68.3} & \textbf{61.6} & \textbf{75.9} & 61.7 \\
\bottomrule
\end{tabular}
\label{tab:codeact_compared_to_others}
\end{table*}

\subsection{RQ1: How does the CodeACT framework perform across different datasets and models?}

We evaluate the performance and efficiency of the CodeACT framework by training models on two distinct datasets, OSS-Instruct and EVOL-Instruct. The focus is on assessing the framework's impact on CodeLlama and DeepSeek-Coder models, using only 40\% of the available training data to highlight efficiency gains without compromising model accuracy.

The results presented in Table \ref{tab:codeact_performance} demonstrate the effectiveness of the CodeACT framework in enhancing both efficiency and performance metrics across various models and datasets. For the OSS-Instruct dataset, the CodeACT framework significantly reduces training times and peak GPU memory usage for both CodeLlama and DeepSeek-Coder models. Specifically, the CodeLlama-7B model trained with CodeACT shows a 71\% reduction in training time (from 220 minutes to 63 minutes) and a 48\% decrease in peak GPU memory usage (from 64.28 GB to 33.21 GB), while also achieving improved performance on the HumanEval and HumanEval+ benchmarks. Similarly, the CodeLlama-13B model exhibits a 70\% reduction in training time and a 14\% reduction in peak GPU memory usage, with enhanced performance on the HumanEval benchmark. The DeepSeek-Coder-6.7B model, despite its smaller size, also benefits from the CodeACT framework, achieving a 71\% reduction in training time and a 45\% reduction in peak GPU memory usage, with the highest scores on most benchmarks.

For the EVOL-Instruct dataset, the CodeACT framework continues to exhibit efficiency gains. The CodeLlama-7B model trained with CodeACT reduces training time by 77\% and peak GPU memory usage by 24\%, with performance improvements on the MBPP+ benchmark. The CodeLlama-13B model achieves a 75\% reduction in training time and a 15\% reduction in peak GPU memory usage, along with performance gains on the HumanEval benchmark. The DeepSeek-Coder-6.7B model achieves a 78\% reduction in training time and a 27\% reduction in peak GPU memory usage, with notable performance improvements on the HumanEval and HumanEval+ benchmarks.

These results suggest that the CodeACT framework not only enhances training efficiency but also maintains or improves model performance across different datasets and models. This indicates that the framework's ability to leverage a smaller subset of training data effectively contributes to both computational savings and performance optimization.

\subsection{RQ2: How does the performance of models trained with CodeACT compare to other models?}

We evaluate the performance of models trained with the CodeACT framework relative to other state-of-the-art models, including both closed-source and open-source variants. Table \ref{tab:codeact_compared_to_others} presents a detailed comparison across various benchmarks.

The closed-source models, such as Claude-3-Opus and GPT-4-Turbo, demonstrate superior performance across most benchmarks, with GPT-4-Turbo achieving the highest scores on HumanEval and HumanEval+ (85.4\% and 81.7\%, respectively). Claude-3-Opus leads on MBPP and MBPP+ benchmarks (89.4\% and 73.3\%, respectively). Despite these impressive results, our focus is primarily on comparing the performance of open-source models with those trained with the CodeACT framework.

Our results show that models fine-tuned with CodeACT exhibit substantial performance improvements compared to their respective base models. For instance, our CodeACT-CL-13B achieves notable gains, scoring 64.0\% on HumanEval and 55.5\% on HumanEval+, significantly outperforming its base model, CodeLlama-13B. Similarly, our CodeACT-DS-6.7B also excels by achieving 68.3\% on HumanEval and 61.6\% on HumanEval+, surpassing the performance of other open-source models with similar or larger parameter sizes, including Magicoder-DS-6.7B and WizardCoder-SC-15B.

Additionally, our models demonstrate impressive efficiency in data utilization. Despite using fewer data samples, models trained with CodeACT achieve better performance compared to models that utilize a larger dataset. For example, the CodeACT-DS-6.7B model, trained on 30K samples, outperforms Magicoder-DS-6.7B, which was trained on 75K samples. This highlights the efficacy of the CDAS method in selecting high-quality, influential data that enhances model performance. Moreover, the performance of CodeACT-DS-6.7B rivals that of closed-source models like Gemini-Pro-1.0, demonstrating that the gap between open-source and closed-source models can be significantly reduced using our framework.

Overall, the experimental results highlight CodeACT's capability to significantly enhance Code LLMs. By providing an efficient and effective training methodology through optimized data selection and training processes, CodeACT enables open-source models to achieve or even surpass the performance of larger, more resource-intensive models.

\subsection{RQ3: Why use K-Means for selecting diverse data?}
\label{subsec:RQ3}

\begin{table}[ht]
\renewcommand{\arraystretch}{1.2}
\centering
\caption{This table presents a comparison of various diverse data selection algorithms and their impact on the performance of the DeepSeek-Coder-Base-6.7B model using the OSS-Instruct dataset.}
\begin{tabular}{lccc}
\toprule
\textbf{Method}& \textbf{Sampling Time} & \textbf{HumanEval} & \textbf{HumanEval+} \\ 
\midrule
K-Center Greedy & 209 min & 58.5\% & 54.3\% \\ 
Graph Density & 36 min & 63.6\% & 57.9\% \\ 
K-Means & 0.2 min & 63.4\% & 57.9\% \\ 
\bottomrule
\end{tabular}
\label{tab:diverse_sampling_methods}
\end{table}

To investigate why K-Means is preferred for selecting diverse data in the CDAS method, we conduct experiments using DeepSeek-Coder-Base-6.7B model on the OSS-Instruct dataset. In this study, we compare three algorithms: K-Center Greedy \cite{sener2018active}, Graph Density \cite{ebert2012ralf} and K-Means. All algorithms are evaluated with a sampling rate set to 40\%. The results are summarized in Table \ref{tab:diverse_sampling_methods}.

The results indicate that K-Means achieves a near-optimal balance between sampling efficiency and model performance. Specifically, K-Means requires only 0.2 minutes for sampling, significantly lower than both K-Center Greedy (209 minutes) and Graph Density (36 minutes). This remarkable efficiency can be attributed to K-Means' algorithmic simplicity and its ability to converge quickly. Despite its rapid sampling time, K-Means maintains competitive performance with 63.4\% and 57.9\% pass rates on HumanEval and HumanEval+, respectively, which are on par with the outcomes yielded by the more time-intensive Graph Density algorithm.

Both K-Center Greedy and Graph Density demonstrate limitations in efficiency compared to K-Means. K-Center Greedy, while producing reasonable results with a 58.5\% pass rate on HumanEval and 54.3\% on HumanEval+, is extremely inefficient with a sampling time of 209 minutes. This inefficiency stems from its iterative nature, continually seeking the point that maximizes the minimum distance to any already chosen point. Graph Density, though more efficient than K-Center Greedy, still requires 36 minutes for sampling. It achieves the highest pass rate of 63.6\% on HumanEval and ties with K-Means at 57.9\% on HumanEval+, but the marginal performance gains do not justify the significantly longer sampling time. Both methods become computationally demanding as dataset size increases, with K-Center Greedy's approach becoming prohibitively slow and Graph Density requiring pairwise distance calculations that scale quadratically with the number of data points. These characteristics limit their applicability in real-world scenarios where time constraints and large-scale datasets are common.

In conclusion, the superior efficiency and comparable performance of K-Means make it the optimal choice for selecting diverse data. Its simplicity and effectiveness in clustering data points based on similarity ensure a diverse dataset without the computational overhead associated with the other algorithms. This comprehensive approach enables the model to generalize better and perform more effectively on coding tasks, validating K-Means as a superior sampling strategy for optimizing model training.

\subsection{RQ4: How should the sampling rate for CDAS be set?}

\begin{figure}
    \centering
    \includegraphics[width=1\linewidth]{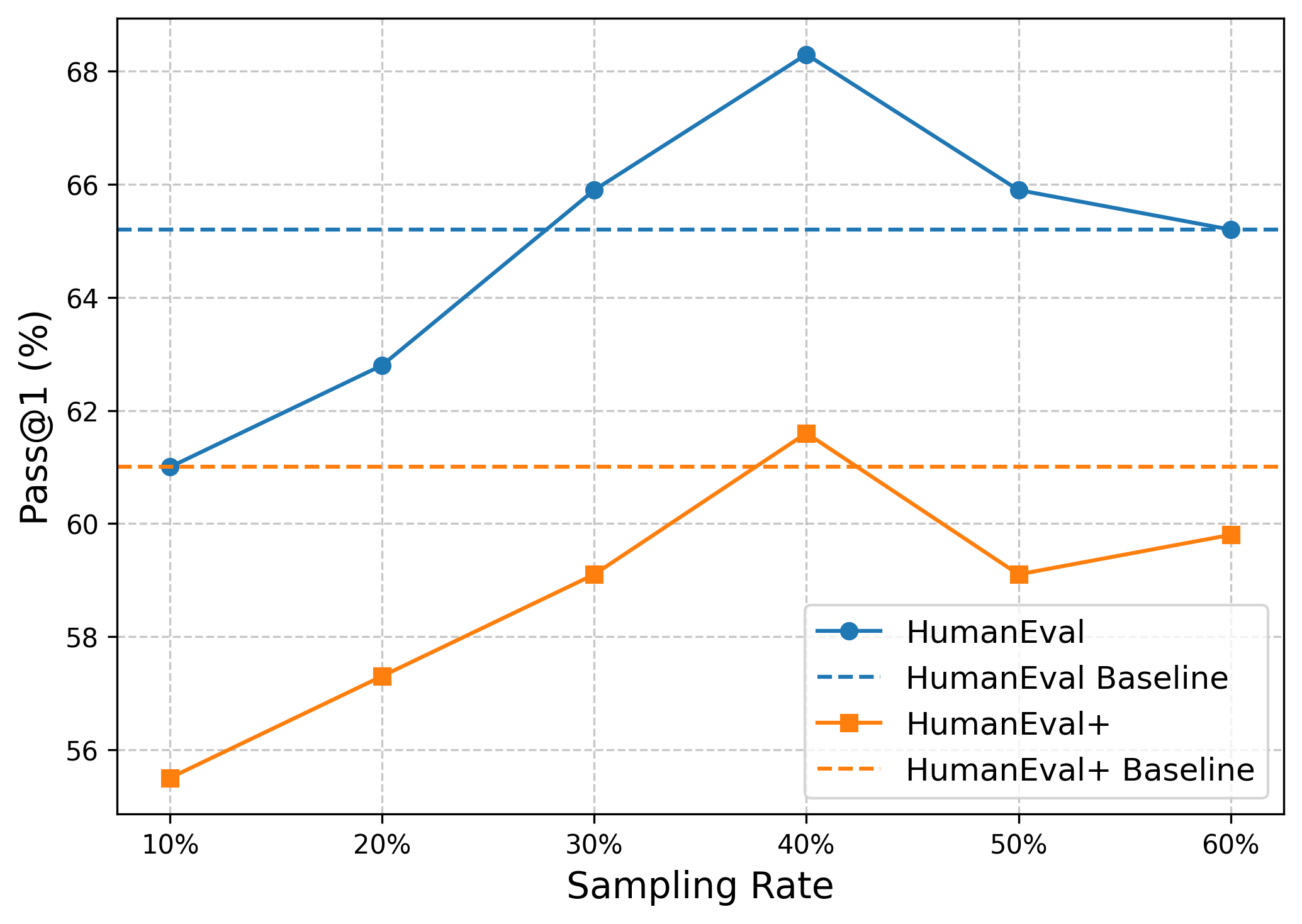}
    \caption{Comparison of sampling rates and their impact on the performance of the DeepSeek-Coder-Base-6.7B model using the OSS-Instruct dataset.}
    \label{fig:sampling_rate_comparison}
\end{figure}

To determine the optimal sampling rate for the CDAS method, we conduct experiments using the DeepSeek-Coder-Base-6.7B model on the OSS-Instruct dataset. The results, shown in Figure \ref{fig:sampling_rate_comparison}, compare the performance of different sampling rates (from 10\% to 60\%) against the baseline (100\% data) on the HumanEval and HumanEval+ benchmarks.

The baseline performance, with full data, achieves 65.2\% on HumanEval and 61.0\% on HumanEval+. When reducing the data to 10\%, the performance drops significantly to 61.0\% on HumanEval and 55.5\% on HumanEval+. Increasing the sampling rate to 20\% improves the performance to 62.8\% on HumanEval and 57.3\% on HumanEval+. At a 30\% sampling rate, the performance further improves to 65.9\% on HumanEval and 59.1\% on HumanEval+. Notably, with a sampling rate of 40\%, the model achieves the highest performance, surpassing the baseline with scores of 68.3\% on HumanEval and 61.6\% on HumanEval+.

However, increasing the sampling rate beyond 40\% results in a decline in performance. At a 50\% sampling rate, the model's performance decreases to 65.9\% on HumanEval and 59.1\% on HumanEval+. Similarly, a 60\% sampling rate sees further declines to 65.2\% on HumanEval. These results suggest that overly large sampling rates may include more redundant or less informative data, which does not contribute to and may even hinder model performance.

The experimental results indicate that a sampling rate of 40\% not only maintains the model's performance but actually improves it compared to using the entire dataset. This improvement can be attributed to the CDAS method's ability to effectively select diverse and complex data, which enhances the model's generalization capabilities and training efficiency. Additionally, using 40\% of the data significantly reduces the computational resources and training time required, making it a more efficient choice.

In conclusion, based on the observed performance gains and efficiency improvements, a 40\% sampling rate is determined to be the optimal setting for the CDAS method. This rate provides a balanced trade-off between maintaining high model performance and reducing computational overhead, thereby validating its selection as the final sampling rate for our experiments.

\subsection{RQ5: How does CDAS compare to other sampling methods in terms of performance?}

\begin{figure}
    \centering
    \includegraphics[width=1\linewidth]{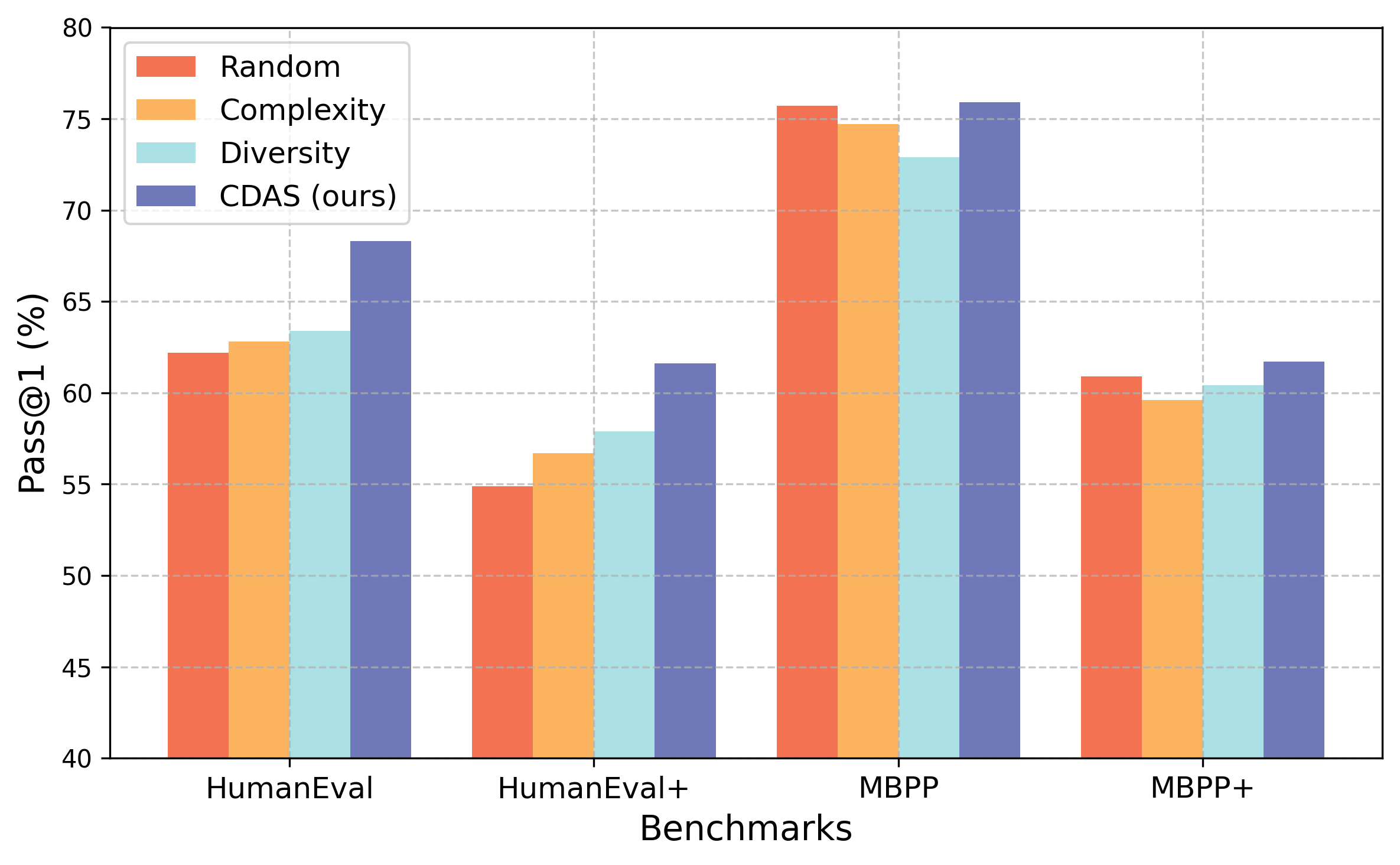}
    \caption{Comparison of sampling methods and their impact on the performance of the DeepSeek-Coder-Base-6.7B model using the OSS-Instruct dataset.}
    \label{fig:sampling_method_comparison}
\end{figure}

To investigate the effectiveness of different sampling methods, we conduct experiments using the DeepSeek-Coder-Base-6.7B model on the OSS-Instruct dataset. The sampling techniques examined include random sampling, complexity sampling, diversity sampling, and our innovative CDAS method. Complexity sampling involves selecting the top m\% of data based on IFD scores. Diversity sampling employs K-Means clustering to randomly extract m\% of data from each cluster. All strategies are evaluated with a sampling rate set to 40\%. The results are summarized in Figure \ref{fig:sampling_method_comparison}.

The results demonstrate that CDAS outperforms the other methods across all benchmarks, achieving the highest scores. Notably, the performance improvement of CDAS over other sampling methods highlights the effectiveness of integrating both complexity and diversity in the data selection process. Random sampling and diversity sampling yield relatively lower performance, indicating that simply ensuring data variety without considering complexity is insufficient. Similarly, complexity sampling alone does not achieve optimal results, underscoring the importance of a balanced approach that considers both the complexity and diversity of training samples.

By considering both complexity and diversity, CDAS ensures that the selected data is not only challenging but also representative of a broad range of programming scenarios. This comprehensive approach enables the model to generalize better and perform more effectively on diverse coding tasks, thus validating CDAS as a superior sampling method for optimizing model training. The synergistic effect of combining complexity and diversity in CDAS leads to a more robust and versatile model, capable of handling a wide array of programming challenges with improved accuracy and efficiency.

\section{Discussion}

\subsection{Threats to Validity}

\textbf{Scope of Model Sizes.} The current study has primarily evaluated the CodeACT framework on models with 7B and 13B parameters. While the results demonstrate significant improvements in both performance and efficiency, it is crucial to validate the framework's effectiveness on larger-scale models. Testing CodeACT on models with greater parameter sizes, such as those with 30B or more parameters, would provide deeper insights into its scalability and generalizability. Evaluating the framework on larger models will also help ascertain whether the observed benefits in smaller models hold true in more complex architectures.

\textbf{Scope of Code-Related Tasks.} The current study has focused exclusively on code generation tasks. However, the efficacy of the CodeACT framework should extend to a wider array of code-related tasks, such as bug fixing\cite{deepfix, bugswarm, PyTER, debugbench} and code summarization\cite{TL-CodeSum, codesearchnet, octopack}. Future research should aim to investigate the applicability and effectiveness of CodeACT in these additional domains to provide a more comprehensive assessment of its capabilities. Evaluating CodeACT on bug fixing tasks would examine its capacity to identify and correct errors in code, thereby enhancing code reliability. Code summarization tasks, which involve generating concise descriptions of code functionality, would further demonstrate the framework's potential to aid in code comprehension and documentation. Extending the evaluation to these diverse tasks would offer a holistic view of CodeACT's utility in the broader software engineering landscape.

\subsection{Limitation of CodeACT}

While the CodeACT framework has demonstrated effectiveness in selecting code data by considering both complexity and diversity, there are inherent limitations that need to be addressed. One significant limitation is the challenge of ensuring the correctness of complex code. Although the CDAS algorithm has proven to be effective, selecting complex data often brings about issues related to the correctness and reliability of the code samples. Complex code, by its nature, is more prone to errors and inconsistencies, which can adversely affect the quality of training and the performance of the resulting models.

Ensuring the correctness of complex code data is a crucial aspect that CodeACT currently does not fully address. The inclusion of erroneous or unreliable code can lead to suboptimal model training, ultimately impacting the model's ability to generate accurate and functional code. Future research should focus on integrating mechanisms to validate and ensure the correctness of the selected code data. This could involve incorporating automated testing, static analysis, or leveraging additional LLMs to verify the accuracy and functionality of the complex code samples before they are used for training.

By addressing this limitation, we aim to enhance the robustness and reliability of the CodeACT framework. Ensuring the correctness of complex code will not only improve the quality of the training data but also contribute to the overall effectiveness of the model in real-world coding tasks. This aspect will be a key focus in our future research efforts.

\section{Related Work}

\subsection{Base Code LLMs}
The advancement of LLMs has significantly impacted various domains, including code generation and understanding. Closed-source models, such as GPT-4\cite{gpt-4}, have consistently ranked highly on mainstream evaluation metrics, demonstrating superior performance in code-related tasks. These models leverage extensive resources and proprietary data, resulting in a performance gap between them and their open-source counterparts.

To bridge this gap and democratize access to advanced coding capabilities, several open-source models have been developed. Notable among these are CodeLlama\cite{codellama}, DeepSeek-Coder\cite{deepseek-coder}, and CodeGemma\cite{codegemma}. CodeLlama, derived from the Llama 2\cite{llama2} architecture, is specifically fine-tuned for code generation tasks and has shown competitive results compared to closed-source models. These open-source models have significantly propelled the field of code generation forward. They offer robust alternatives to closed-source models, promoting innovation and collaboration within the research community. The continuous development and refinement of these models are crucial for narrowing the performance disparity and advancing the state-of-the-art in code-related applications.

\subsection{Data Generation}
A significant area of research focuses on generating instructional data to fine-tune base LLMs. Self-Instruct\cite{self-instruct} is one such method that refines weaker student models by using strong teacher models to generate synthetic instructions. This approach leverages the expertise of advanced LLMs to produce diverse and complex instructional data, which in turn helps in training more effective student models. Evol-Instruct\cite{wizardcoder} takes this a step further by iteratively increasing the complexity of the coding instructions. This method involves evolving instructions to be more challenging over multiple iterations, thereby improving the model's ability to handle complex tasks. Another notable approach is OSS-Instruct\cite{magicoder}, which generates realistic coding problems based on open-source code snippets. By extracting real-world code segments from repositories such as GitHub, OSS-Instruct prompts LLMs to create relevant and practical coding challenges, ensuring the generated data closely mirrors actual programming scenarios.

These methodologies typically utilize more powerful LLMs, such as GPT-4, to generate vast amounts of synthetic data. While these approaches address the issue of data quantity, they often overlook data quality. Ensuring the relevance, diversity, and complexity of the generated data remains a critical challenge for further enhancing model performance.

\subsection{Data Selection}

The process of manual data curation is not only costly but also susceptible to subjective bias, rendering the development of automated data selection methods critically important. Current automated data selection methods are primarily divided into two categories: those that rely on external models for data selection and those that do not.

In the realm of methods dependent on external models, for instance, AlpaGasus\cite{alpagasus} utilizes meticulously crafted prompt templates to leverage ChatGPT for scoring data quality. InsTag\cite{instag} employs ChatGPT to obtain detailed labels for each data instruction, assessing the complexity and diversity of the data based on these labels. LIFT\cite{lift} generates a diverse set of instructions using GPT-4 to augment the dataset, followed by vectorization and selection of subsets based on row variables, ultimately utilizing GPT-4 for multi-dimensional scoring of the data. While these methods have demonstrated efficacy in handling large-scale datasets, they also incur significant economic costs.

In contrast, methods independent of external models, such as DQ\cite{dq}, integrate techniques of data distillation and coreset selection\cite{iyer2021submodular}. The core technology involves defining a gain function to iteratively partition the dataset and select representative samples, thereby maximizing data diversity. Cherry LLM\cite{ifd} introduces the Instruction-Following Difficulty score, determined by comparing the cross-entropy loss of model-generated responses with and without instructions. A high IFD score implies that the model struggles to accurately align answers with instructions, reflecting the complexity of the instructions.

Despite the progress made in automated data curation, methods specifically tailored for code data selection remain notably absent from the literature.

\section{Conclusion}

In this paper, we introduce the CodeACT framework, designed to optimize the training of Code LLMs by addressing both data quality and computational efficiency. CodeACT integrates the CDAS method for selecting complex and diverse data, and the Dynamic Pack padding strategy to minimize padding tokens and reduce resource consumption. Our experimental results demonstrate that CodeACT-DeepSeek-Coder-6.7B, fine-tuned on only 40\% of the EVOL-Instruct data, achieves a significant performance increase on HumanEval by 8.6\%, a reduction in training time by 78\%, and a decrease in peak GPU memory usage by 27\%. These findings validate the effectiveness of the CodeACT framework in improving both the performance and efficiency of Code LLMs.

Future work should focus on ensuring the correctness of complex code data. While CDAS effectively selects influential data, it will be crucial to integrate mechanisms for validating and improving the accuracy of complex instructions. This will further strengthen the CodeACT framework, making it more effective in handling diverse and complex coding tasks.

\bibliography{references}{}
\bibliographystyle{IEEEtran}

\end{document}